\pdfoutput=1
\documentclass[11pt]{article}
\usepackage[margin=1in]{geometry}
\usepackage{graphicx}
\usepackage[utf8]{inputenc} 
\usepackage[T1]{fontenc}    
\usepackage{amsfonts}      
\usepackage{microtype}      
\usepackage{amsmath}
\usepackage{csquotes}
\usepackage{mhequ}
\usepackage{algorithm}
\usepackage{algorithmic,color,url}
\usepackage{mathptmx}
\usepackage{amsthm}
\usepackage[numbers]{natbib}

\begin{document}
\title{Variational Gaussian approximation of the Kushner optimal filter \footnote{This document is the preprint version of the article published for the International Conference on Geometric Science of Information, page 395-404, 2023. }}
%
%
\author{%
  Marc Lambert  \\
  DGA/CATOD, Centre d'Analyse Technico-Op\'erationelle de D\'efense\\
  \& INRIA\\
  \texttt{marc-h.lambert@intradef.gouv.fr} 
  \and
  Silvère Bonnabel \\
MINES ParisTech, PSL University, Center for robotics\\
  \texttt{silvere.bonnabel@mines-paristech.fr}  
  \and
  Francis Bach \\
  INRIA - Ecole Normale Sup\'erieure - PSL Research university  \\
  \texttt{francis.bach@inria.fr} 
}
 \date{}
\maketitle              
\begin{abstract}
In estimation theory, the Kushner equation provides the evolution of the probability density of the state of a dynamical system given continuous-time observations. Building upon our recent work, we propose a new way to approximate the solution of the Kushner equation through tractable variational Gaussian approximations of two proximal losses associated with the propagation and Bayesian update of the probability density. The first is a proximal loss based on the Wasserstein metric and the second is a proximal loss based on the Fisher metric. The solution to this last proximal loss is given by implicit updates on the mean and covariance that we proposed earlier. These two variational updates can be fused and shown to satisfy  a set of stochastic differential equations on the Gaussian's mean and covariance matrix. This Gaussian flow is consistent with the Kalman-Bucy and Riccati flows in the linear case and generalize them in the nonlinear one.
\end{abstract}

\section{Introduction}
We consider the general filtering problem where we aim to estimate the state $x_t$ of a continuous-time stochastic system given noisy observations $y_t$. If the state follows a Langevin dynamic $f=-\nabla V$ with $V$ a potential function and the observations occur continuously in time, the problem can be described by two stochastic differential equations (SDE) on $x_t$ and $z_t$, where $z_t$ is related to the observation by the equation $dz_t=y_tdt$:
\begin{align}
&dx_t=-\nabla V(x_t)dt+\sqrt{2 \varepsilon}d\beta \label{SDEx}\\
&dz_t=h(x_t)dt+\sqrt{R}d\eta. \label{SDEz}
\end{align}
$\beta$ and $\eta$ are independent Wiener processes and $Q=2 \varepsilon \mathbb{I}$ and $R$ play the role of covariance matrices of the associated diffusion processes. Many dynamical systems can be rewritten in the Langevin canonical form \eqref{SDEx}, see for instance  \cite{Halder2017}. In essence  \eqref{SDEz} means $``y_t=h(x_t)+\text{noise}"$, but one has to resort to \eqref{SDEz} to avoid problems related to infinitely many observations. The optimal Bayesian filter corresponds to the conditional probability $p_t$ of the state at time $t$ given all past observations. This probability satisfies the Kushner equations which can be split into two parts:
\begin{align} 
dp_t=\mathcal{L} (p_t) dt+d\mathcal{H}(p_t), \label{Kushner}
\end{align}
where $\mathcal{L}$ is defined by the Fokker-Planck partial differential equation (PDE) 
\begin{align} 
\mathcal{L}(p_t)={\rm{div} }[\nabla V p_t] + \varepsilon \Delta p_t,\label{FP}
\end{align}
whereas the second term corresponds to the Kushner stochastic PDE (SPDE): $$d\mathcal{H}(p_t)=(h-\mathbb{E}_{p_t}[h])^TR^{-1}(dz_t-\mathbb{E}_{p_t}[h]dt)p_t,$$
where $\mathbb{E}_{p_t}[h]:=\int h(x) p_t(x)dx$ and stochasticity comes from $dz_t$. These equations cannot be solved in the general case, and we must resort to approximation. In this paper, we consider variational Gaussian approximation, which consists in searching  for the  Gaussian distribution $q_t$ closest to the optimal one $p_t$ for a particular variational loss. Two variational losses are well suited for our problem. 

Jordan-Kinderlehrer-Otto (JKO)  \cite{Jordan98}  showed that the following   proximal scheme: 
\begin{align}
\text{argmin} ~\mathcal{L}^{\delta t}(p)= \text{argmin} ~\left[ KL \left(p \Big|\Big| \pi \right)+\frac{1}{2\delta t } d^2_w(p_t,p) \right],\quad\text{(JKO)}\label{JKO}
\end{align} is related to the Fokker-Planck (FP) equation associated to \eqref{SDEx} where we denote its stationary distribution $\pi\propto\exp(-V/\varepsilon)$. Indeed, iterating this proximal algorithm yields a curve being solution to FP as $\delta t\to 0$. It is referred to as variational since it  is an optimization problem over the function $p$, and it  involves the Kullback-Leibler divergence defined by $KL(p||\pi)=\int p \log \frac{p}{\pi},$ and the Wasserstein (or optimal transport) distance $d^2_w(p_t,p)$ \cite{Ambrosio2005}.

The variational loss associated to the Kushner PDE is the Laugesen-Mehta-Meyn-Raginsky (LMMR) proximal scheme \cite{Laugesen2014} defined  by: 
\begin{align}
\text{argmin} ~\mathcal{H}^{\delta t} (p)=\text{argmin} ~\left[\mathbb{E}_p \frac{1 }{2}  ||\delta z_t-h(x) \delta t||^2_{(R \delta t)^{-1}}+ KL(p||p_t) \right],~\text{(LMMR) }\label{LMMR}
\end{align}
where $\delta z_t:=z_{t+\delta t}-z_t$ comes from the Euler-Marayama discretization of the observation SDE: $\delta z_t=h(x_t)\delta t+\sqrt{R} \delta \eta$ such that $p(\delta z_t|x_t)=\mathcal{N}(h(x_t) \delta t,R \delta t)$. 

For small  $\delta t$ those schemes generate a sequence of probability distributions that   converge to the solutions of the corresponding PDE in the limit  $\delta t \rightarrow 0$. We see the KLs in both schemes play a different role, though. In \eqref{JKO}, the proximal scheme shows that the solution to the FP equation follows a gradient of the KL to the stationary distribution $\pi$. This gradient is computed with respect to the Wasserstein metric. In \eqref{LMMR}, the proximal scheme defines a gradient over the state prediction $p$ of the expected prediction error. This gradient is computed in the sense of the metric defined by the KL around its null value, which may be related to the Fisher metric.



To approximate the solutions, we propose to constrain them to lie in the space of Gaussian distributions. That can be done by constraining in the proximal schemes the general distribution $p_t$ to be a Gaussian distribution $q_t=\mathcal N(\mu,P)$. 
The proximal problems become finite-dimensional and boil down to minimizing $\mathcal{L}^{\delta t}$ and $\mathcal{H}^{\delta t}$ over $(\mu,P)$. The Gaussian approximation of the JKO scheme yields in the limit a set of ODEs on $\mu$ and $P$ as shown in \cite{Lambert22c}. In this paper, we extend these results showing the Gaussian solution to the LMMR scheme corresponds to the R-VGA solution \cite{Lambert22a} which yields in the limit a set of SDEs on $\mu$ and $P$. Moreover, using a two-step approach, we can fuse the two Gaussian solutions to approximate the Kushner equation \eqref{Kushner}. As shall be shown presently, we find the following SDEs for $\mu$ and $P$:
\begin{align} \label{CKF}
&\boxed{\begin{aligned} 
& \textbf{The fully continuous-time variational Kalman filter} \\
&d \mu_t= b_tdt + P_tdC_t \\
&dP_t = A_tP_tdt+P_tA_t^Tdt+\frac{1}{2}  dH_tP_t+\frac{1}{2}  P_tdH_t^T +2 \varepsilon \mathbb{I} dt  \\
&\text{where } b_t=-\mathbb{E}_{q_t}[\nabla V(x)] ; \quad dC_t=\mathbb{E}_{q_t}[\nabla h(x_t)^T R^{-1}(dz_t-h(x_t)dt)]  \\
&A_t=-\mathbb{E}_{q_t}[\nabla^2 V(x)];\quad dH_t=\mathbb{E}_{q_t}[(x_t-\mu_t)(dz_t-h(x_t)dt)^TR^{-1}\nabla h(x_t)]. 
\end{aligned} }
\end{align}
The  equation for $P_t$ can be seen as a generalization of the Riccati equation in the nonlinear case. Indeed, if we replace $V$ and $h$ with linear functions, the ODE on $P_t$ matches the Riccati equations and we recover the Kalman-Bucy filter, known to solve exactly the Kushner equations. 

This paper is organized as follows: Section \ref{secRW} is dedicated to related works on the approximation of the optimal nonlinear filter. In Section \ref{secLMMR}  we derive the variational Gaussian approximation of the LMMR scheme. In Section \ref{secJKO}  we recall the variational Gaussian approximation of the JKO scheme proposed in our previous work. In Section \ref{secKushner} we combine these two results to obtain the Continuous Variational Kalman filter equations and  show the equivalence with the Kalman-Bucy filter in the linear case. 

\section{Related works} \label{secRW}
In 1967, Kushner proposed a Gaussian assumed density filter to solve his PDE \cite{Kushner1967}. This filter is derived by keeping only the first two moments of $p_t$ in \eqref{Kushner} which can be computed in closed form using the Ito formula. These moments involve integrals under the unknown distribution $p_t$ and the heuristic is to integrate them rather on the current Gaussian approximation $q_t$ leading to a recursive scheme. A more rigorous way to do this approximation was proposed later \cite{Hanzon1991,Brigo99} with the projected filter. In this approach, the solution of the Kushner PDE is projected onto the tangent space to Gaussian distributions equipped with the Fisher information metric.  This leads to ODEs  that are quite different from \eqref{CKF}. A third approach is to linearize the stochastic dynamic process to obtain a McKean Vlasov process that allows  for   Gaussian propagation \cite[Sec 4.1.1]{Lambert22b}. The connexion between approximated SDEs and projected filters was analyzed in detail earlier in \cite{Brigo97}. \\
The latter approach is the one explored in the current paper, i.e., considering   proximal schemes associated with the Kushner PDE  where we constrain the solution to be  Gaussian. It is equivalent to projecting the exact gradient flow onto the tangent space of the manifold of  Gaussian distributions. This approach is preferred since it exhibits the problem’s geometric structure and allows convergence guarantees to be proven. Approximation of gradient flows is an active field and several recent papers have followed this direction: the connexion between the propagation part of the Gaussian assumed density filter and the variational JKO scheme \cite{Jordan98} was recently studied  \cite{Lambert22c}; the connexion between the update part of the Gaussian assumed density filter and the variational LMMR scheme \cite{Laugesen2014} was studied in \cite{Halder2018} where a connexion with a gradient flow was first established but limited to the linear case. To the best of our knowledge, the variational approximation of the LMMR scheme in the nonlinear case has never been addressed. 
The various ways to obtain the ODEs \eqref{CKF} lead to nice connexions between geometric projection, constrained optimization, and statistical linearization. These different approaches are illustrated in Figure \ref{fig1} which addresses only the approximation of dynamics \eqref{SDEx} 
  without measurements (i.e., propagation only) for which all  methods prove equivalent. 
\begin{figure}
\begin{tabular}{ccc}
Distribution $p$ && Gaussian approx. $q(\mu,P)$\\
&&\\
  $dx_t=-\nabla V(x_t) dt$ & \normalsize \rm{SDE linearization \cite{Lambert22b}} &  $dx_t=A(t)(x_t-\mu_t)dt+b(t)dt$\\
$+\sqrt{2\varepsilon}d\beta$  &&$+\sqrt{2\varepsilon}d\beta$\\
\framebox{nonlinear SDE process } &$\Rightarrow$   & \framebox{ McKean–Vlasov process} \\
  &&\\
  &&\\
$\frac{\partial p}{\partial t} = {\rm{div}} (\nabla V p) +\varepsilon \Delta p$ & &    $\dot{\mu_t}=b(t)$\\ 
&\normalsize Riemanian projection \cite{Hanzon1991} &   $\dot{P_t}=A(t)P_t+P_tA(t)^T+2\varepsilon \mathbb{I} $\\
 \framebox{Fokker-Planck } & $\Rightarrow$ & \framebox{ Variational Gaussian flow} \\
  &&\\
  &&\\
$KL(p||\pi) + \frac{1}{2\delta t }  d_w(p,p_{t})^2 $ \vspace{0.05cm}   & \normalsize \rm{constrained optim. \cite{Lambert22c} }   & $KL(q||\pi) + \frac{1}{2\delta t }  d_{bw}(q,q_{t})^2$ \\
\framebox{proximal JKO} & $\Rightarrow$   &  \framebox{proximal Bures-JKO} \\
  &&\\
  &&\\
  $\delta p= -\nabla_{w} KL (p||\pi)$ & \normalsize \rm{gradient projection \cite{Lambert22c} } &   $\delta q= -\nabla_{bw} KL (q||\pi)$   \\
\framebox{W2 gradient flow} &  $\Rightarrow$ & \framebox{Bures-W2 gradient flow} 
 \end{tabular}
 \vspace{0.2cm}
 $$\text{where}  \quad A(t)=-\mathbb{E}_{q_t}[\nabla^2 V(x)] ;\quad \quad b(t)=-\mathbb{E}_{q_t}[\nabla V(x)]$$
\caption{Various equivalent approaches for Gaussian approximation of the SDE \eqref{SDEx}.
We denote $\pi\propto\exp(-V/\varepsilon)$  as the stationary distribution of the associated Fokker-Planck equation.
The left column presents equivalent definitions of $p$ whereas the right column corresponds to the approximated solution $q$ in the space of Gaussian distributions. $d_w$ denotes the Wasserstein distance whereas $d_{bw}$ denotes the Bures-Wasserstein distance, which is its restriction to the subset of Gaussian distributions. At the last row, the tangent vector  $\delta p$, respectively (resp. $\delta q$) and the gradient $\nabla_{w}$ (resp. $\nabla_{bw}$) are defined with respect to the Wasserstein metric space  of distribution $\left(\mathcal{P}(\mathbb{R}^d),d_w^2\right)$ (resp. the Bures-Wasserstein metric space of Gaussians $\left(\mathcal{N}(\mathbb{R}^d),d_{bw}^2\right)$). These geometries are briefly explained in Section \ref{W2geom}. 
}
 \label{fig1}
\end{figure}

\section{Variational Gaussian approximation of the LMMR proximal} \label{secLMMR}

In this section, we compute the closest Gaussian solution to the LMMR problem \eqref{LMMR}. The corresponding Gaussian flow is closely related to natural gradient descent used in information geometry. This flow approximates the Kushner optimal filter when the state is static. In the next sections, we will generalize this result to a dynamic state. 

\subsection{The recursive variational Gaussian approximation}
The proximal LMMR problem \eqref{LMMR} where we constrain the solution $q$ to be a Gaussian reads (given $q_t$ a current Gaussian distribution at time $t$):
\begin{align} 
q_{t+\delta t}&=\underset{q \in \mathcal{N}(\mu,P)}{\arg \min} \quad \mathbb{E}_q \frac{1}{2}  ||\delta z_t-h(x)\delta t||^2_{(R\delta t)^{-1}}+ KL(q||q_t) \\
&=\underset{q \in \mathcal{N}(\mu,P)}{\arg \min} \quad - \int q(x) \log p(\delta z_{t}|x)dx  + KL(q||q_t)\\
&=\underset{q \in \mathcal{N}(\mu,P)}{\arg \min} \quad KL\left(q \Big|\Big| \frac{1}{Z}p(\delta z_{t}|x)q_t\right),
\label{LMMRq}\end{align}
where we have introduced a normalization constant $Z$ which does not change the problem. \\
Eq \eqref{LMMRq} falls into the framework of variational Gaussian approximation (R-VGA) \cite{Lambert22a}. The solution satisfies the following updates  \cite[Theorem 1]{Lambert22a}:
\begin{align*}
&\mu_{t+\delta t}=\mu_t+P_t\mathbb{E}_{q_{t+\delta t}}[\nabla_x \log p(\delta z_{t}|x)] \\
&P^{-1}_{t+\delta t}=P_t^{-1} - \mathbb{E}_{q_{t+\delta t}}[\nabla_x^2 \log p(\delta z_{t}|x)],
\end{align*}
where the expectations are under the Gaussian $q_{t+\delta t} \sim \mathcal{N}(\mu_{t+\delta t},P_{t+\delta t})$ making the updates implict. In the linear case, that is, if we take  $h(x)=Hx$, these updates are equivalent to the online Newton algorithm \cite[Theorem 2]{Lambert22a}. Computing the Hessian $\nabla_x^2 \log p$ can be avoided using integration by part: 
$$P^{-1}_{t+\delta t}=P_t^{-1} - P^{-1}_{t+\delta t}\mathbb{E}_{q_{t+\delta t}}[(x-\mu_{t+\delta t})\nabla_x \log p(\delta z_{t}|x)^T]$$
By rearranging the terms and using that $P$ is symmetric (see \cite[Sec 4.2]{Lambert22b}) we can let appear an update on
the covariance:
$$ P_{t+\delta t}=P_t+\frac{1}{2}  \mathbb{E}_{q_{t+\delta t}}[(x-\mu_t)\nabla_x \log p(\delta z_{t}|x)^T]P_t+\frac{1}{2}  P_t \mathbb{E}_{q_{t+\delta t}}[\nabla_x \log p(\delta z_{t}|x)(x-\mu_t)^T]. $$
Finally, using that $\nabla_x \log p(\delta z_{t}|x)=\nabla h(x)^T R^{-1}(\delta z_t-h(x)\delta t)$ we obtain:
\begin{align} 
&\mu_{t+\delta t}=\mu_t+ P_t\delta C_t, \quad P_{t+\delta t}=P_t+\frac{1}{2}  \delta H_tP_t+\frac{1}{2}  P_t \delta H_t^T,  \label{LMMRupdates} \\
&\text{ where: } \nonumber\\
&\delta C_t=\mathbb{E}_{q_{t+\delta t}}[\nabla h(x)^T R^{-1}(\delta z_{t}-h(x)\delta t)] \nonumber \\
&\delta H_t=\mathbb{E}_{q_{t+\delta t}}[(x-\mu_t)(\delta z_{t}-h(x)\delta t)^TR^{-1}\nabla h(x)]. \nonumber
\end{align} 
Letting $\delta t\to 0$, we obtain the following SDE in the sense of Ito: 
\begin{align} 
&d\mu_t= P_td C_t, \quad dP_t=\frac{1}{2}  d H_tP_t+\frac{1}{2}  P_t d H_t^T,   \label{SDELMMR}
\end{align} 
where it shall be noted that $dH_t$ is non-deterministic owing to $dz_t$. Since the LMMR scheme has been proven to converge to the solution of the Kushner SPDE \cite{Laugesen2014}, this SDE describes the best Gaussian approximation of the optimal filter when the state is static. 


\subsection{Information geometry interpretation}
We show here how the LMMR proximal scheme is related to the Fisher information geometry in the general case.
Let’s consider  a family of  densities:
 $S=\Big\{ p(.|\theta); \theta \in \Theta; \Theta \subseteq \mathbb{R}^m \Big\}$ and let $$F(\theta)=\int \nabla_\theta \log p(x|\theta)  \nabla_\theta \log p(x|\theta)^T p(x|\theta) dx,$$ be the Fisher information matrix, where $\theta$ regroups  all the parameters.
If we consider now the proximal LMMR on $S$, and if we use the second-order Taylor expansion of the KL divergence between these two distributions, we have:
$$KL(p(x|\theta)||p(x|\theta_t))=\frac{1}{2}  (\theta-\theta_t)^T F(\theta_t) (\theta-\theta_t) + o((\theta-\theta_t)^2).$$
Rather than minimizing the proximal LMMR scheme \eqref{LMMR} in the infinite space of distributions, we now search the minimum in the finite space of parameters:  
\begin{align*}
\theta_{t+\delta t}=\underset{\theta \in \Theta}{\arg \min} \quad \mathbb{E}_{p(x|\theta)}\left[ \frac{1}{2}  ||\delta z_t-h(x)\delta t||^2_{(R\delta t)^{-1}}\right]+ \frac{1}{2}  (\theta-\theta_t)^T F(\theta_t) (\theta-\theta_t).
\end{align*}
Considering that the minimum must cancel the gradient of the above proximal loss, we obtain: 
\begin{align}
&0 = \nabla_{\theta} \left(  \mathbb{E}_{ p(x|\theta)}\left[ \frac{1}{2}  ||\delta z_t-h(x)\delta t||^2_{(R\delta t)^{-1}} \right] \right)\big \vert_{\theta_{t+\delta t}} + F(\theta_t)(\theta_{t+\delta t}-\theta_t)\nonumber \\
&\theta_{t+\delta t} = \theta_t -F(\theta_t)^{-1} \nabla_{\theta} \left(\frac{1}{2}   \mathbb{E}_{ p(x|\theta)}\left[ ||\delta z_t-h(x)\delta t||^2_{(R\delta t)^{-1}} \right] \right)\big \vert_{\theta_{t+\delta t}},\label{GradFlow0}
\end{align}
which corresponds to a gradient descent of the averaged stochastic likelihood:
\begin{align}
\theta_{t+\delta t} &= \theta_t - F(\theta_t)^{-1} \nabla_{\theta} \mathbb{E}_{p(x|\theta)}[ -\log p(\delta z_t|x)]\big \vert_{\theta_{t+\delta t}}.\label{GradFlow}
\end{align}
Remarkably, the optimal filer equations with a static state $x$ are given by an implicit Bayesian variant of the natural gradient descent \cite{Amari1998}. Indeed here $x$ plays the role of the parameter of the likelihood distribution. The original natural gradient should be a descent with the gradient $-F(x_t)^{-1}\nabla_x \log p(\delta z_t|x)\big \vert_{x_t}$. 


\section{Variational Gaussian approximation of the JKO proximal} \label{secJKO}

The canonical Langevin form \eqref{SDEx} assumes that the drift term $f=-\nabla V$ derives from a potential $V$. This potential has a physical meaning in filtering (consider a gravity field for example). The evolution of the state in the filter mimics the true evolution of the physical system. It’s not the case in statistical physics, where the potential is constructed such that $V=-\log \pi$ where $\pi$ is the asymptotic distribution of a variable $x$ which doesn’t correspond to a physical system. We used this property in our previous work \cite{Lambert22c} and simulated a dynamic to approximate the target $\pi$ with a Gaussian distribution. Here we do not want to estimate a distribution but to propagate a Gaussian through the nonlinear physical dynamic \eqref{SDEx}. 

\subsection{The Bures-JKO proximal}

The proximal JKO problem \eqref{JKO} where we constrained the solution $q$ to be a Gaussian distribution writes:
  \begin{align*}
 &\underset{q \in \mathcal{N}(\mu,P)}{\min} \quad  KL\left(q \Big|\Big|\pi\right) + \frac{1}{2\delta t } d_{bw}(q,q_t)^2,
 \end{align*}
where  $d_{bw}(q,q_t)$ is the Bures distance between two Gaussians given by: 
  \begin{align}
  d_{bw}(q,q_t)=||\mu-\mu_t||^2 +\mathcal{B}^2(P,P_t), \label{BuresDist}
   \end{align} 
 where  $\mathcal{B}^2(P,P_t)=\mathrm{Tr}(P+ P_t - 2  (P^{\frac{1}{2}} P_tP^{\frac{1}{2}})^{\frac{1}{2}})$ is the squared Bures metric \cite{Bures69}, which has a derivative available in closed form. After some computation \cite[Appendix A]{Lambert22c} we can obtain implicit equations that the parameters of the optimal Gaussian solution $q$  must satisfy:
  \begin{align} 
 & \mu_{t+\delta t}=\mu_t-\delta t. \mathbb{E}_{q_{t+\delta t}}[\nabla V(x)] \nonumber \\
 & P_{t+\delta t}=P_t-\delta t. \mathbb{E}_{q_{t+\delta t}}[\nabla^2 V(x)]P_t - \delta t.  P_t\mathbb{E}_{q_{t+\delta t}}[\nabla^2 V(x)]^T+ 2\varepsilon \delta t.  \mathbb{I}, \label{JKOupdates}
  \end{align}
  and at the limit $\delta t  \rightarrow 0$, we obtain the following ODEs:
  \begin{align} 
 & \dot{\mu_t}=-\mathbb{E}_{q_t}[\nabla V(x)]:=b_t \label{JKOode}  \\
 &  \dot{P_t}=A_tP_t+P_tA_t^T+2\varepsilon \mathbb{I} \quad \text{ where } A_t:=-\mathbb{E}_{q_t}[\nabla^2 V(x)].\nonumber
  \end{align}

\subsection{Wasserstein geometry interpretation} \label{W2geom}
The Wasserstein geometry is defined by the metric space of measure endowed with the Wasserstein distance $\left(\mathcal{P}(\mathbb{R}^d),d_w^2\right)$. The definition of a tangent vector in this space is tedious because the measure $\mu$ must satisfy the conservation of mass $\int \mu(x)dx =1$. To handle this constraint we can use the continuity equation. This equation allows to represent any regular curves of measures with a continuous flow along a vector field $v_t \in \mathbb{L}^2$. It is closely related to the Fokker-Planck equation as we show now (see \cite{Ambrosio2005} for more details). 
The JKO proximal scheme \eqref{JKO} gives a sequence of distribution that satisfies at the limit the Fokker-Planck equation \eqref{FP}, this equation rewrites as follows:
\begin{align}
\dot p_t &=\nabla.(\nabla V p_t) + \varepsilon \nabla. \nabla p_t=\nabla.(\nabla V p_t) + \varepsilon \nabla. (p_t \nabla \log p_t) \nonumber\\
&=\nabla.(p_t (\nabla V + \varepsilon \nabla\log p_t))=-{\rm{div}}(p_tv_t), \label{continuityEq}
\end{align}
which is a continuity equation where $v_t \in \mathbb{L}^2(\mathbb{R}^d)$ plays the role of the tangent vector $\delta p_t$ along the path $p_t$ and satisfies: $$v_t=-\nabla V - \varepsilon \nabla\log p_t=-\nabla_w KL(p_t||\pi),$$ with $\pi \propto \exp(-V/\varepsilon)$. The last equality comes from variational calculus in the measure space: the Wasserstein gradient of a functional $F$ is given by the Euclidian gradient of the first variation $\nabla_w F(\rho)=\nabla \delta F(\rho)$, see \cite[Chapter 10]{Ambrosio2005}.

Let's sum up what's going on: starting from a stochastic state $x_t$ following the Langevin dynamic \eqref{SDEx} with drift $-\nabla V$, we have rewritten the Fokker-Planck equation which describes the evolution of the density $p(x_t)$ as a continuity equation \eqref{continuityEq} where the diffusion term has disappeared. At this continuity equation correspond a deterministic ODE $\dot x_t= -\nabla_w KL(p_t||\pi)$. It's a nice property of the Wasserstein geometry where PDE can be described by a continuity equation that corresponds to a simple  gradient flow.  


Following the same track, the sequence of Gaussian distributions satisfying the ODE \eqref{JKOode} correspond to a Wasserstein gradient flow given by the continuity equation: $\dot q_t =-div(q_t w_t),$ where $w_t=-\nabla_{bw} KL(q_t||\pi)$ is now a gradient with respect to the Bures-Wasserstein distance \eqref{BuresDist}, see \cite[Appendix B3]{Lambert22c} for the analytical expression of this gradient.

\section{Variational Gaussian approximation of the Kushner optimal filter} \label{secKushner}

We have tackled the two proximal problems independently but how to solve them jointly? The simplest method to do so is to alternate between propagation through dynamics \eqref{SDEx} for a small time $\delta t$, and Bayesian update through LMMR in the light of the accumulated observations $\delta z_t$, and let $\delta t\to 0.$ This is what we do presently. 

\subsection{The continuous variational Kalman filter}\label{secCKF}
Consider one step of the Euler–Maruyama method with length $\delta t$ of  SDEs \eqref{SDEx} and \eqref{SDEz}. As the Wiener processes $\beta$ and $\eta$ are  independent, we may write:
$$p(x_t,y_{t+\delta t},x_{t+\delta t})=p(y_{t+\delta t}|x_{t+\delta t},x_t)p(x_{t+\delta t},x_t)=p(y_{t+\delta t}|x_{t+\delta t})p(x_{t+\delta t}|x_t),$$
 denoting $y_{t+\delta t}=\delta z_t$. In other words, we can solve the proximal LMMR  update equation \eqref{LMMRq} using as prior $q_t(x)=\mathcal{N}(\mu_{t+\delta t|t}, P_{t+\delta t|t})$,    the solution of the proximal JKO. The LMMR/R-VGA discrete-time equations \eqref{LMMRupdates} then become:
\begin{align*} 
&\mu_{t+\delta t}=\mu_{t+\delta t|t}+P_{t+\delta t|t}\delta C_t\\
&P_{t+\delta t}=P_{t+\delta t|t}+ \frac{1}{2} \delta H_t P_{t+\delta t|t}+  \frac{1}{2} P_{t+\delta t|t}\delta H_t^T. 
\end{align*} 
Replacing $\mu_{t+\delta t|t}$ and $P_{t+\delta t|t}$ by their expressions as the solutions to the JKO scheme \eqref{JKOupdates} and putting in a residual all the terms in $\delta t^2$, we obtain: 
\begin{align*}
\mu_{t+\delta t}&=\mu_t+\delta t b_t+ P_t \delta C_t\\
P_{t+\delta t}&=P_t+\delta t A_tP_t + \delta t  P_tA_t^T+ \delta t  2\varepsilon \mathbb{I}+\frac{1}{2} \delta H_t P_t+\frac{1}{2} P_t\delta H_t^T + O(\delta t^2). 
\end{align*}
By Ito calculus, we obtain the continuous variational Kalman updates \eqref{CKF}. 

\subsection{The Kalman-Bucy filter as a particular case}
Let us consider the linear case where the SDEs \eqref{SDEx} and \eqref{SDEz} rewrite:
$$dx_t=Fx_tdt+\sqrt{2 \varepsilon}d\beta, \quad dz_t=Gx_tdt+\sqrt{R}d\eta.$$
The various expectations that appear in the proposed filter \eqref{CKF} apply either to quantities being independent of $x_t$ or being linear  or quadratic in $x_t$, yielding
\begin{align*}
& d\mu_t= F\mu_tdt + P_tG^TR^{-1}(dz_t-G\mu_tdt)\\
&\frac{d}{dt}P_t = FP_t+P_tF^T-P_tG^TR^{-1}GP_t+2\varepsilon \mathbb{I}.
\end{align*}
We see we exactly  recover the celebrated Kalman-Bucy filter. 

\section*{Conclusion}
We have approximated the Kushner optimal filter by a Gaussian filter based on variational approximations related to the JKO and LMMR proximal discrete schemes related to the Wasserstein and Fisher geometry respectively. 
As the dynamic and observation processes are assumed independent, we can mix the two variational solutions to form a set of SDEs on the Gaussian parameters generalizing the Riccati equations associated to the linear systems. In the linear case, the proposed filter boils down to  the Kalman-Bucy optimal filter. It is still unclear, though,  which global variational loss is minimized by the optimal filter. 

\subsection*{Acknowledgements}
This work was funded by the French Defence procurement agency (DGA) and by the French government under the management of Agence Nationale de la Recherche as part of the “Investissements d’avenir” program, reference ANR-19-P3IA-0001(PRAIRIE 3IA Institute). We also acknowledge support from the European Research Council (grant SEQUOIA 724063).

\bibliographystyle{plain}
\bibliography{NonLinearFiltering}

\end{document}